\documentclass[conference]{IEEEtran}
\IEEEoverridecommandlockouts
\usepackage{cite}
\usepackage{amsmath,amssymb,amsfonts}
\usepackage{graphicx}
\usepackage{textcomp}
\usepackage{xcolor}
\def\BibTeX{{\rm B\kern-.05em{\sc i\kern-.025em b}\kern-.08em
    T\kern-.1667em\lower.7ex\hbox{E}\kern-.125emX}}

\usepackage{graphicx}

\usepackage{amssymb}
\usepackage{mathtools}

\usepackage{diagbox}
\newcommand{\tabincell}[2]{\begin{tabular}{@{}#1@{}}#2\end{tabular}}
\usepackage{booktabs}

\usepackage{algorithm,algpseudocode}

\usepackage{xcolor}

\usepackage{multirow}

\algnewcommand{\algorithmicforeach}{\textbf{for each}}
\algdef{SE}[FOR]{ForEach}{EndForEach}[1]
  {\algorithmicforeach\ #1\ \algorithmicdo}
  {\algorithmicend\ \algorithmicforeach}
\usepackage{algpseudocode}

\begin{document}

\title{PPSEBM: An Energy-Based Model with Progressive Parameter Selection for \\ Continual Learning\\
}

\author{\IEEEauthorblockN{1\textsuperscript{st} Xiaodi Li}
\IEEEauthorblockA{\textit{Department of Artificial Intelligence and Informatics} \\
\textit{Mayo Clinic}\\
Rochester, USA \\
li.xiaodi@mayo.edu}
\and
\IEEEauthorblockN{2\textsuperscript{nd} Dingcheng Li}
\IEEEauthorblockA{\textit{Vertex AI} \\
\textit{Google}\\
Kirkland, USA \\
dingchengl@gmail.com}
\and
\IEEEauthorblockN{3\textsuperscript{rd} Rujun Gao}
\IEEEauthorblockA{\textit{Department of Mechanical Engineering} \\
\textit{Texas A\&M University}\\
College Station, USA \\
grj1214@tamu.edu}
\and
\IEEEauthorblockN{4\textsuperscript{th} Mahmoud Zamani}
\IEEEauthorblockA{\textit{Department of Electrical and Computer Engineering} \\
\textit{The University of Texas at Dallas}\\
Richardson, USA \\
mxz173130@utdallas.edu}
\and
\IEEEauthorblockN{5\textsuperscript{th} Feng Mi}
\IEEEauthorblockA{\textit{Computer Science Department} \\
\textit{The University of Texas at Dallas}\\
Richardson, USA \\
Feng.Mi@utdallas.edu}
\and
\IEEEauthorblockN{6\textsuperscript{th} Latifur Khan}
\IEEEauthorblockA{\textit{Computer Science Department} \\
\textit{The University of Texas at Dallas}\\
Richardson, USA \\
lkhan@utdallas.edu}
}

\maketitle

\begin{abstract}
Continual learning remains a fundamental challenge in machine learning, requiring models to learn from a stream of tasks without forgetting previously acquired knowledge. A major obstacle in this setting is catastrophic forgetting, where performance on earlier tasks degrades as new tasks are learned. In this paper, we introduce PPSEBM, a novel framework that integrates an Energy-Based Model (EBM) with Progressive Parameter Selection (PPS) to effectively address catastrophic forgetting in continual learning for natural language processing tasks. In PPSEBM, progressive parameter selection allocates distinct, task-specific parameters for each new task, while the EBM generates representative pseudo-samples from prior tasks. These generated samples actively inform and guide the parameter selection process, enhancing the model’s ability to retain past knowledge while adapting to new tasks. Experimental results on diverse NLP benchmarks demonstrate that PPSEBM outperforms state-of-the-art continual learning methods, offering a promising and robust solution to mitigate catastrophic forgetting.
\end{abstract}


\section{Introduction}
Imagine a real-world scenario in which a language model is deployed in a virtual assistant that continually learns and adapts to new natural language processing (NLP) tasks, such as sentiment analysis, question answering, dialogue generation, and information extraction, without forgetting critical knowledge from prior tasks. Continual learning is essential in this context to ensure the model can efficiently handle a diverse range of user queries and maintain consistent performance across tasks. However, a key challenge arises when the model encounters new tasks, which can lead to a decline in performance on previously mastered tasks—a problem known as catastrophic forgetting. In response to this critical issue, our paper introduces PPSEBM, an innovative solution that seamlessly integrates an Energy-Based Model (EBM) with Progressive Parameter Selection (PPS).

An energy-based model is a type of machine learning model that learns to associate a scalar energy value with each possible input data point \cite{lecun2006tutorial}. The fundamental principle behind energy-based models lies in the notion that lower energy values are assigned to data points that are more likely to be true or belong to the target distribution, while higher energy values are given to data points that are less likely or deviate from the desired distribution \cite{hinton2006reducing}. Additionally, the energy-based model's generative nature allows it to generate new data points that resemble the characteristics of the training data, making it useful for data generation tasks as well \cite{lecun2006tutorial,bengio2009learning,lecun2015deep}.

Parameter Selection is derived from Prompt Tuning, a powerful and innovative approach to fine-tuning language models \cite{brown2020language,holtzman2019curious}. Unlike traditional fine-tuning, where models are trained on large, labeled datasets, prompt tuning focuses on generating high-quality prompts, which are short input strings or questions that guide the model's responses \cite{petroni2019language}. In our work, the prompt refers to the random selection of model parameters \cite{razdaibiedina2023progressive}.

In this paper, we propose an innovative approach PPSEBM, to tackle the challenge of catastrophic forgetting in continual learning for language models. For each new task encountered, the model progressively learns distinct parameters. These parameters are then concatenated with all previously learned parameters. Crucially, we employ an energy-based model to generate additional training data from previous tasks, and this data not only reinforces prior knowledge but also actively informs the parameter selection process. By providing representative samples from past tasks, the EBM guides PPS to focus parameter updates on task-relevant components, thereby enhancing stability and adaptability. Through this tight integration of generative replay and selective parameter adaptation, the model significantly mitigates catastrophic forgetting.

Our paper presents three primary contributions: (1) We introduce an innovative method that combines progressive parameter selection with energy-based models for continual learning, which effectively addresses catastrophic forgetting; (2) Our approach includes the ability of the energy-based model to generate additional training samples from previous tasks and further guiding the parameter selection process; (3) Our method achieves state-of-the-art performance, demonstrating its effectiveness and superiority.

\section{Related Works}

\subsection{Energy-based Models}
The paper \cite{lecun2006tutorial} introduces Energy-Based Models (EBMs) to represent probability density functions $p(x)$ using an energy function $E(x):R^{D}\rightarrow R$. These models assign low energy values to realistic points and high energy values to unrealistic points, providing advantages in simplicity and stability during single model training. EBMs can reduce parameters through shared features and avoid prior assumptions that could lead to bottlenecks \cite{du2019implicit}. Recent developments enable data space EBMs, as described in \cite{xie2016theory,nijkamp2019learning}, to model complex dependencies in various data types like images, videos, 3D shapes, and point clouds. Latent space EBMs, introduced in \cite{pang2020learning}, enhance model expressivity for generating text, images, and trajectories. Another work \cite{zhang2021learning} proposes an energy-based informative prior to capture the latent data space, which is more expressive than other priors. Similarly, \cite{li2022energy} demonstrate the potential of EBMs in preventing catastrophic forgetting in continual learning, achieving strong performance in class-incremental scenarios. Unlike our method, where EBM serves as an outer layer for various tasks, Li et al. solely focus on using EBM as the entire model and only concentrate on the classification tasks in the image domain while ours focus on a diverse range of tasks in the language domain. Recently, \cite{li2025lsebmcl} proposes using EBM to generate pseudo-samples in continual learning and achieving satisfying performance.

\subsection{Parameter Selection}
Recent work shows parameter selection \cite{lester2021power,jung2023generating,kim2025open} offers an efficient alternative to full finetuning by training soft prompts \cite{razdaibiedina2023progressive} while keeping pretrained weights $\theta$ fixed. The selected parameters $P$ with dedicated weights $\theta_p$ are added to the input $x$ and updated independently. Different from \cite{razdaibiedina2023progressive}, we use only the parameter loss for backpropagation and fine-tune the language model parameters throughout. Several advanced methods—DualPrompt \cite{wang2022dualprompt}, SAPT \cite{zhao2024sapt}, and L2P \cite{wang2022learning}—extend these ideas for continual learning, mainly in computer vision (CV). The difference between our work and theirs is that our work focuses on the NLP domain, and most of them focus on the CV domain, and some of these (e.g., DualPrompt, L2P) are relatively early works from 2022.

\subsection{Continual Learning}
Continual learning, also known as life-long learning or incremental learning, involves learning tasks sequentially. Deep neural networks implementing continual learning face the challenge of \textit{catastrophic forgetting} \cite{shin2017continual}. Three families of continual learning approaches are Replay methods, Regularization-based methods, and Parameter isolation methods \cite{zhou2023revisiting,zhou2024continual}. Replay methods store or generate samples, such as A-GEM \cite{chaudhry2018efficient} and LAMOL \cite{sun2019lamol}. Regularization-based methods add a regularization term to consolidate previous knowledge, including LwF \cite{li2017learning}, MAS \cite{aljundi2018memory}, and LPC \cite{li2022lpc}. Parameter isolation methods allocate different model parameters to each task, such as MoCL \cite{wang2024rehearsal} and SAPT \cite{zhao2024sapt}. Our method is both a replay and parameter isolation method.

\section{Methodology}
\begin{figure*}[ht]
    \centering
    \includegraphics[width=6.3in]{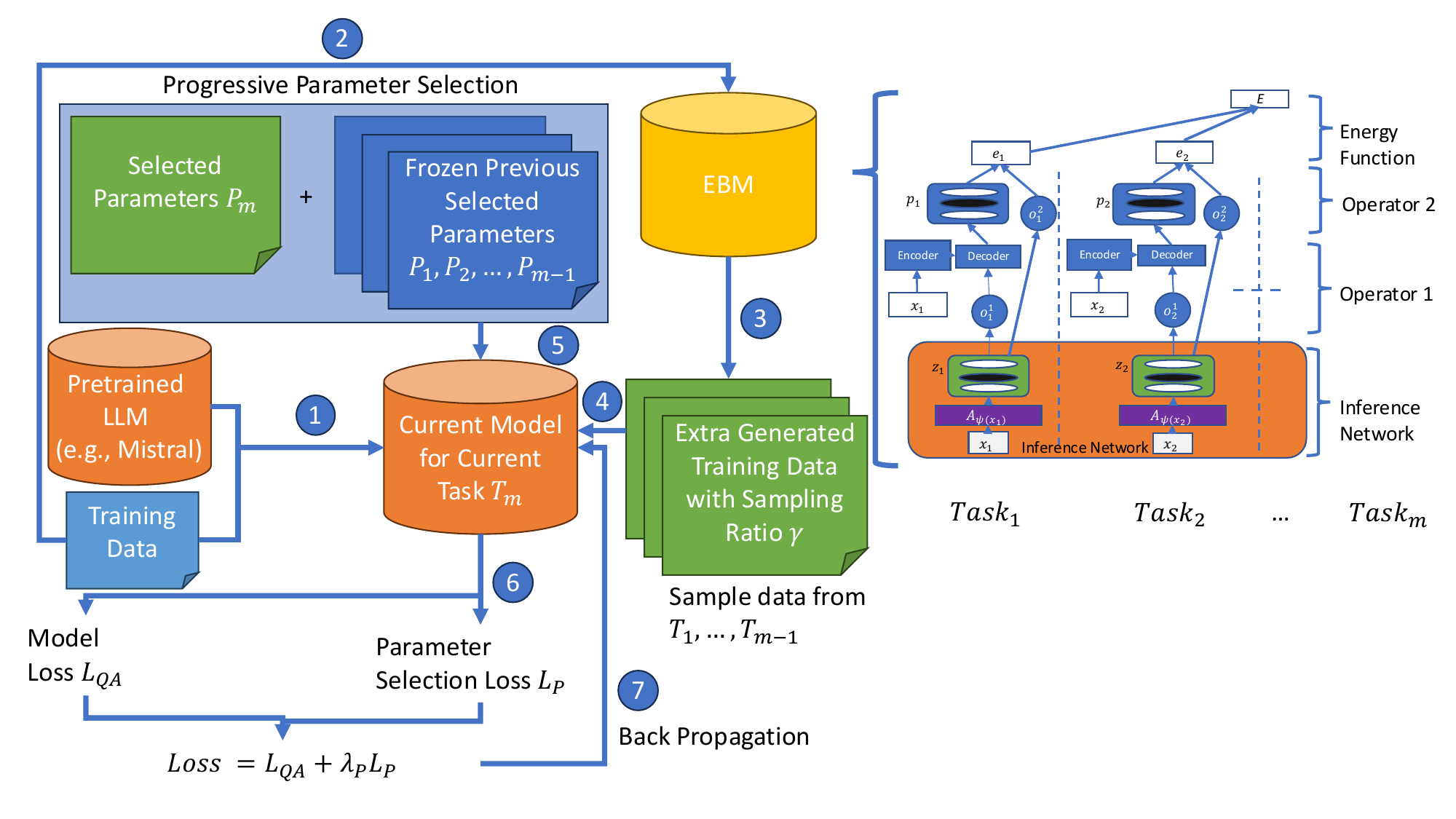}
    \caption{The Overview of PPSEBM Framework.}
    \label{fig:Overview}
\end{figure*}
In this section, we present our novel approach to address catastrophic forgetting in continual learning using the Energy-Based Model (EBM) and Progressive Parameter Selection (PPS) (Figure \ref{fig:Overview}). The EBM generates supplementary training samples from previous tasks $T_1, ..., T_{m-1}$ and integrates them into the training of the new task $T_m$ to preserve knowledge. Meanwhile, the PPS process concatenates past selected parameters (a small portion of random model parameters, e.g. 10 rows of parameters) with the current one, enabling adaptive parameter selection for individual tasks. Our approach involves the following steps as shown in Figure \ref{fig:Overview}: (1) We initiate the process with a pre-trained base model, specifically the Large Language Model (LLM) Mistral 7B. \cite{jiang2023mistral}. (2) We train one EBM using available training data in the sequence order of different tasks for all tasks simultaneously during the model training. (3) The trained EBM generates additional training data (sampling ratio $\gamma$) for previous tasks iteratively using the same sequence order. (4) The extra generated small training data is combined with the current training data in the task $P_m$ to train a model by updating parameters. (5) We calculate the selected parameters loss. For this, we initialize the parameters by selecting random parameters from the current model, and the current parameters are progressively initialized by concatenating them with previously frozen selected parameters from past tasks during model training. (6) Model training involves combining the model loss and selected parameters loss to form the final loss. (7) We update the model parameters through backpropagation using gradient descent based on the calculated loss. 

\subsection{Energy-based Model}
\subsubsection{Inference Network}
We follow decaNLP \cite{mccann2018natural} and preprocess datasets into a unified Question Answering (QA) format: $<x, y> \in D$, where $x$ represents the Question and $y$ denotes the Answer. The dataset $D$ is used for training and covers diverse tasks like Question Answering (QA), Natural Language Inference (NLI), Sentiment Analysis (SA), Semantic Role Labeling (SRL), etc.

We introduce an inference network denoted as $A_{\Psi}(x)$, also known as the energy-based inference network (Figure \ref{fig:Overview}). It is parameterized by $\Psi$ and trained to achieve the following goal:

\begin{equation}
A_{\Psi}(x) \approx \underset{y\in Y_{R}(x)}{argmin} E_{\Theta}(x, y)
\end{equation}

The inference network parameters $\Psi$ are trained as follows ($\Theta$ is the parameters of the EBM):

\begin{equation}
\widehat{\Psi} = \underset{\Psi}{argmin}\sum_{<x, y>\in D} E_{\Theta}(x, A_{\Psi}(x))
\end{equation}

\subsubsection{Operator 1}
We use two operators, $o^{1}$ and $o^{2}$, to map logits $z_{m}$ into distributions for the energy function. $o^{1}$ is responsible for modulating how the logits $z_{m}$ from the inference network are fed to the decoder input slots in the energy function, as depicted in the right-hand side of Figure \ref{fig:Overview}. It handles the operation of feeding inference network outputs into the decoder input slots in the energy.

\subsubsection{Operator 2}
We use operator $o^{2}$ to determine how the distribution $p_{\Theta}(\cdot| ...)$ is used to compute the log probability of $y$. It handles the operation of computing the energy on the output. The local energy term is written as:

\begin{equation}
e_{m}(x,y) = -o^{2}(z_{m})\log p_{\Theta}(\cdot|o^{1}(z_{m}), x_{m})
\end{equation}

Our objective is to minimize this energy function with respect to the variable $z^{(t)}$ in our inference networks. The SX (softmax) operation is selected for $o^1$ and $o^2$. Here, $o(z) = softmax(z)$. $m \in [1, M]$ means the index of the current task and $M$ means the total number of tasks.

The Encoder and Decoder used in our EBM are Gaussian Mixture Variational Autoencoders including an EncoderRNN as the encoder and a DecoderRNN as the decoder.

\subsubsection{Energy Function}
We train an EBM as an outer-generator. We train an EBM simultaneously with the model training, one after another. After training each task, the trained EBM model generates samples from the current task, stores them in a directory concatenated with previously generated samples, and adds them to the new upcoming tasks (steps 3 and 4 of Figure \ref{fig:Overview}). The energy function is defined as follows:

\begin{equation}
E_{\Theta}(x, y) = \sum_{m=1}^{M}e_{m}(x, y)
\end{equation}

\noindent where $\Theta$ is the parameters of the EBM, and $e_{m}(x, y)=-\log p(y_{m}|x_{m})$.

For each task, we have:

\begin{equation}
    p_{\theta}(x, z) = p_{\alpha}(z)p_{\beta}(x|z)
\end{equation}

\noindent where $p_{\alpha}(z)$ is the prior model with parameters $\alpha$, $z$ is a latent dense continuous vector, and $p_{\beta}(x|z)$ is given by a generative model parameterized with $\beta$. 

For language modeling, $x\in R^{D}$ and $x = g_{\beta}(z) + \epsilon$, where $\epsilon\sim \mathcal{N}(0, \sigma^{2}I_{D})$, so that $p_{\beta}(x|z)\sim \mathcal{N}(g_{\beta}(z), \sigma^{2}I_{D})$. $g_{\beta}(z)$ maps the prior model $p_{\alpha}$ to be closer to the data distribution.

For text modeling, let $x=(x^{(t)}, t=1, ... T)$ where each $x^{(t)}$ is a token and $p_{\beta}(x|z)$ is an autoregressive model as shown in equation \ref{equ_autoregressive}: 

\begin{equation}\label{equ_autoregressive}
    p_{\beta}(x|z)=\prod_{t}^{T}p_{\beta}(x^{(t)}|x^{(1)}, ..., x^{(t-1)}, z)
\end{equation}

\noindent which is parameterized by a recurrent network with parameters $\beta$. $t \in [1, T]$ means the current learning iteration and $T$ means the total number of learning iterations.

The EBM prior of $z$, $p_{\alpha}(z)$, is an energy-based correction of an isotropic Gaussian reference distribution $p_{0}(z)$:

\begin{equation}
\begin{aligned}
    p_{\alpha}(z) = \frac{1}{Z(\alpha)}\exp(f_{\alpha}(z))p_{0}(z)
\end{aligned}
\end{equation}

\noindent where $p_0(z)$ is a known reference distribution, assumed to be isotropic Gaussian in this paper. $f_\alpha(z)$ is the negative energy and is parameterized by a small multi-layer perceptron with parameter $\alpha$. $Z(\alpha) = \int \exp(f_\alpha(z))p_0(z)dz = E_{p_0}[\exp(f_\alpha(z))]$ is the normalizing constant or partition function.

The marginal distribution is $p_{\theta}(x) = \int p_{\theta}(x, z)dz = \int p_{\alpha}(z) p_{\beta}(x|z)dz$. The posterior distribution is $p_{\theta}(z|x) = p_{\theta}(x, z) / p_{\theta}(x) = p_{\alpha}(z) p_{\beta}(x|z) / p_{\theta}(x)$.

We draw samples from the EBM prior $p_{\alpha}(z)$ and the posterior distribution $p_{\theta}(z|x)$ using short-run MCMC, Langevin dynamics. We draw samples from the EBM prior:

\begin{equation}
\begin{aligned}
    z_{k + 1} = z_{k} + s\nabla_{z}\log p_{\alpha}(z_{k}) + \sqrt{2s}\epsilon_{k}, \\
    k = 1, ..., K, z_{0}\sim p_{0}(z), \epsilon_{k}\sim \mathcal{N}(0, I_{d})
\end{aligned}
\end{equation}

The short-run Langevin dynamics is always initialized from the fixed initial distribution $p_{0}$, and only
runs a fixed number of $K$ steps, e.g., $K = 20$.

Similarly, we can also draw samples from the posterior distribution $p_{\theta}(z|x)$ by replacing the $p_{\alpha}(z_{k})$ with $p_{\theta}(z_{k}|x)$. The step size of Langevin dynamics is denoted as $s$, and $\nabla_{z}\log p_{\alpha}(z_{k})$ or $\nabla_{z}\log p_{\theta}(z_{k}|x)$ can be efficiently computed by back propagation.

\subsection{Progressive Parameter Selection}
We present our novel approach called Progressive Parameter Selection. In Progressive Parameter Selection, we progressively learn distinct selected parameters $P_{m}$ for each novel task $T_{m}$ encountered during the learning process (as depicted in step 5 Figure \ref{fig:Overview}). For every task $T_{m}$, we train specific parameters $P_{m}$ and concatenate it with all the previously learned selected parameters $P_{i}$, where $i < m$, before prepending them to the input embeddings. During training, we ensure that the parameters $\theta_{P_{m}}$ related to the selected parameters $P_{m}$ are trainable during the learning of task $T_{m}$ and then frozen thereafter. The generated samples by EBM also guide the selection of the parameters. We only retain a limited portion of the selected parameters (e.g., 10) for each task, ensuring that the space usage remains modest in this scenario.

The primary objective in the learning process of task $T_{m}$ (where $m \in \{1,...,M\}$) is to find the optimal selected parameters $\theta_{P_{m}}$ that minimize the negative log probability of training examples, considering both our progressive parameters and the frozen previous selected parameters. The training objective can be defined as follows:

\begin{equation}
\label{equ: prompt}
\begin{aligned}
    L(\theta_{P_{m}}) = -\sum_{x,y\in T_{m}}\log p(y|[P_{m}, ..., P_{1}, x], \\\theta_{p_{1}}, ..., \theta_{P_{m}})
\end{aligned}
\end{equation}

The progressive setup we adopt in Progressive Parameter Selection brings about two significant benefits for efficient continual learning. First, it effectively eliminates the problem of catastrophic forgetting, ensuring that old tasks suffer less forgetting in performance when new tasks are introduced. Second, the separately learned selected parameters for previous tasks facilitate information reuse for future tasks. This phenomenon has been shown in prior research by \cite{vu2021spot}, where selected parameters learned on informative source tasks serve as valuable initializations for downstream tasks, contributing to improved learning efficiency and generalization. The objective is to minimize two types of losses: the QA loss denoted as $L_{QA}$ and the parameter selection loss denoted as $L_{P}$. The overall loss is formulated as $L = L_{QA} + \lambda_{P}L_{P}$, where $\lambda_{P}$ represents the weight of the parameter selection loss.

\subsection{Algorithm}
In this section, we consolidate the various components, including the Inference Network, Operator 1, Operator 2, Energy Function, and Progressive Parameter Selection, into a unified algorithm, as depicted in Algorithm \ref{alg:PPSEBM}.

\begin{algorithm*}
    \centering
    \caption{PPSEBM}
    \begin{algorithmic}[1]
    \State \textbf{given} processed datasets into a QA (Question Answering) format: $<x, y> \in D$, introduce the inference network $A_\Psi(x)$ as an energy-based inference network, employ operators $o^{1}$ and $o^{2}$ to map $z_{m}$ logits, design an EBM layer as a outer-generator, selected paremeters length $p\_len$, learning iterations $T$, learning rate for prior model $\eta_{0}$, learning rate for prior model $\eta_{1}$, initial parameters $\theta_{0}=(\alpha_{0}, \beta_{0})$, observed examples $\{x_{i}\}^{n}_{i=1}$, batch size $b$, number of prior and posterior sampling steps $\{K_{0}, K_{1}\}$, prior and posterior sampling step sizes $\{s_{0}, s_{1}\}$, the total number of tasks $M$, and the index of current task $m$.
    \State \textbf{initialize} timestep $t\gets0, m \gets 0$
    \State
    \State EBM training
    \Repeat
        \Repeat
            \State $A_{\Psi}(x) \approx argmin_{y\in Y_{R}(x)}E_{\Theta}(x, y)$
            \State $\widehat{\Psi} \gets argmin_{\Psi}\sum_{<x, y>\in D}E_{\Theta}(x, A_{\Psi}(x))$ \algorithmiccomment{Inference Network}
            \State $o^1$ modulates the output from the inference network to the energy function's decoder input slots. \algorithmiccomment{Operator 1}
            \State $o^2$ determines how the distribution $p_\Theta(...)$ computes the log probability of $y$. \algorithmiccomment{Operator 2}
            \State $e_{m}(x,y) \gets -o^{2}(z_{m})\log p_{\Theta}(\cdot|o^{1}(z_{m}), x_{m})$ 
            \State $E_{\Theta}(x, y) \gets \sum_{m=1}^{M}e_{m}(x, y)$
            \State $e_{m}(x, y)\gets-\log p(y_{m}|x_{m})$ \algorithmiccomment{Energy Function}
            \State $p_{\theta}(x, z) \gets p_{\alpha}(z)p_{\beta}(x|z)$
            \State $p_{\alpha}(z) \gets \frac{1}{Z(\alpha)}\exp(f_{\alpha}(z))p_{0}(z)$
            \State $Z(\alpha) \gets \int \exp(f_\alpha(z))p_0(z)dz$ \algorithmiccomment{Latent Space Model}
            \State $p_{\beta}(x|z)=\prod_{t}^{T}p_{\beta}(x^{(t)}|x^{(1)}, ..., x^{(t-1)}, z)$ \algorithmiccomment{Model for Text Modeling}
            \State $k \gets 1, ..., K, z_{0}\sim p_{0}(z), \epsilon_{k}\sim \mathcal{N}(0, I_{d})$
            \State $z^{-}_{k + 1} \gets z_{k} + s_{0}\nabla_{z}\log p_{\alpha}(z_{k}) + \sqrt{2s}\epsilon_{k}, K=K_{0}$ \algorithmiccomment{Prior Sampling using MCMC (Langevin dynamics)}
            \State $z^{+}_{k + 1} \gets z_{k} + s_{1}\nabla_{z}\log p_{\theta}(z_{k}|x) + \sqrt{2s}\epsilon_{k}, K=K_{1}$ \algorithmiccomment{Posterior Sampling using MCMC (Langevin dynamics)}
            \State $\alpha_{t+1} \gets \alpha_t + \eta_{0} \frac{1}{b} \sum_{i=1}^b \left( \nabla_{\alpha} f_{\alpha_{t}}(z_i^+) - \nabla_{\alpha} f_{\alpha_{t}}(z_i^-) \right)$ \algorithmiccomment{Learning prior model}
            \State $\beta_{t+1} \gets \beta_t + \eta_{1} \frac{1}{b} \sum_{i=1}^b \nabla_\beta \log p_{\beta_{t}}(x_i|z_i^+)$ \algorithmiccomment{Learning generation model}
            \State $t\gets t + 1$
        \Until{t = T}
        \State $m \gets m + 1$
    \Until{m = M}
    \State \Return optimized parameters $\theta^{M}=(\alpha^{M}, \beta^{M})$
    \State
    \State Upcoming model training
    \Repeat
        \State Extract the generated pseudo samples of previous tasks by the trained EBM
        \State Adding generated previous samples to the current task $<x, y>$
        \State Randomly extract $p\_len$ parameters from the base model as one new selected parameters $P_{m}$
        \State Concatenate selected parameters $P_{m}, ..., P_{1}$
        \State $L(\theta_{P_{m}}) \gets -\sum_{x,y\in T_{m}}\log p(y|[P_{m}, ..., P_{1}, x], \theta_{p_{1}}, ..., \theta_{P_{m}})$ \algorithmiccomment{Calculate the parameter selection loss}
        \State $L = L_{QA} + \lambda_{P}L_{P}$ 
        \State Calculate gradients, do back propagation
    \Until{stopping criterion is met}
    \end{algorithmic}
\label{alg:PPSEBM}
\end{algorithm*}

\section{Experiments}
In this section, we evaluate the performance of our model across a range of tasks. We conduct a comprehensive comparison with eleven different techniques, demonstrating the effectiveness of our approach in addressing the challenge of catastrophic forgetting commonly encountered in continual learning scenarios. To make a fair comparison, all the baselines are retrained on the LLM Mistral 7B. Our work follows task incremental learning, but we generate pseudo-samples across all tasks.
\begin{table*}
\caption{Summary of tasks, datasets, dataset sizes, and their corresponding metrics. As this work uses no development set, only the training and test datasets are shown. nF1 is the normalized version of the F1 score; EM represents an exact match between texts: for text classification, this amounts to accuracy; for WOZ, it is equivalent to dfEM (turn-based dialogue state exact match); for WikiSQL, it is equivalent to lfEM (exact match of logical forms).}
\centering
    \begin{tabular}{llrrr}
         \toprule
         \textbf{Task} & \textbf{Dataset} & \textbf{\# Train} & \textbf{\# Test} & \textbf{Metric} \\
         \midrule
         Question answering & SQuAD 2.0 & 130319 & 11873 & nF1 \\
         Semantic parsing & WikiSQL & 56355 & 15878 & lfEM \\
         Sentiment analysis & SST & 6920 & 1821 & EM \\
         Semantic role labeling & QA-SRL & 6414 & 2201 & nF1 \\
         Goal-oriented dialogue & WOZ & 2536 & 1646 & dsEM \\
         \midrule
         \multirow{5}{*}{Text classification} & AGNews & \multirow{5}{*}{115000} & \multirow{5}{*}{7600} & \multirow{5}{*}{EM} \\
         \multirow{5}{*}{} & Amazon & \multirow{5}{*}{} & \multirow{5}{*}{} & \multirow{5}{*}{} \\
         \multirow{5}{*}{} & DBPedia & \multirow{5}{*}{} & \multirow{5}{*}{} & \multirow{5}{*}{} \\
         \multirow{5}{*}{} & Yahoo & \multirow{5}{*}{} & \multirow{5}{*}{} & \multirow{5}{*}{} \\
         \multirow{5}{*}{} & Yelp & \multirow{5}{*}{} & \multirow{5}{*}{} & \multirow{5}{*}{} \\
         \bottomrule
    \end{tabular}
\label{tab:data}
\end{table*}

\begin{table*}
\caption{Summary of averaged metric scores for different methods under permuted task orders using models at the last epoch of the last task. The Average and Std columns respectively are the average and standard deviation of the averaged scores for each row of the methods. Multitasked learning as an upper bound is shown at the bottom.}\centering
  \resizebox{\textwidth}{30mm}{
  \begin{tabular}{lcccccccc}
    \toprule
    \textbf{Model} & \tabincell{c}{\textbf{SST SRL WOZ}} &
    \tabincell{c}{\textbf{SST WOZ SRL}} &
    \tabincell{c}{\textbf{SRL SST WOZ}} & \tabincell{c}{\textbf{SRL WOZ SST}} & 
    \tabincell{c}{\textbf{WOZ SST SRL}} &
    \tabincell{c}{\textbf{WOZ SRL SST}} & \tabincell{c}{\textbf{Average}} & \tabincell{c}{\textbf{Std}} \\ 
    \midrule
    Fine-tuned & 52.0 & 25.5 & 64.6 & 33.2 & 34.6 & 34.6 & 40.8 & 14.6 \\
    EWC & 51.0 & 50.0 & 66.4 & 37.5 & 44.8 & 40.7 & 48.4 & 10.2 \\
    MAS & 37.3 & 46.2 & 57.3 & 32.3 & 50.3 & 32.2 & 42.6 & 10.3 \\
    GEM & 52.1 & 31.5 & 64.0 & 32.9 & 45.3 & 36.7 & 43.8 & 12.6 \\
    LAMOL $_{GEN}^{0}$ & 47.1 & 38.4 & 57.8 & 39.4 & 45.5 & 46.6 & 45.8 & 7.0 \\
    LAMOL $_{GEN}^{0.05}$ & 81.5 & 79.5 & 74.8 & 73.7 & 70.2 & 75.8 & 75.9 & 4.1 \\
    LAMOL $_{GEN}^{0.2}$ & 80.8 & 81.1 & 81.2 & 80.3 & 78.9 & 81.9 & 80.7 & 1.0 \\
    RVAE-LAMOL $_{GEN}^{0.05}$ & 80.7 & 79.6 & 79.9 & 80.2 & 79.6 & 78.0 & 79.7 & 0.9 \\
    RVAE-LAMOL $_{GEN}^{0.2}$ & 81.5 & 82.3 & 81.5 & 82.0 & 81.1 & 82.8 & 81.9 & 0.6 \\
    LSEBMCL $_{GEN}^{0}$ & 66.4 & 57.7 & 77.7 & 66.8 & 64.6 & 65.6 & 66.5 & 6.4 \\
    LSEBMCL $_{GEN}^{0.05}$ & 82.8 & 81.8 & \textbf{84.4} & 81.9 & 82.7 & 80.5 & 82.4 & 1.3 \\
    LSEBMCL $_{GEN}^{0.2}$ & 83.1 & 82.5 & 82.7 & \textbf{82.4} & 83.7 & 83.2 & 82.9 & \textbf{0.5} \\
    PPSEBM $_{GEN}^{0}$ & 78.7 & 54.6 & 75.6 & 62.8 & 61.4 & 61.6 & 65.8 & 9.3 \\
    PPSEBM $_{GEN}^{0.05}$ & 82.8 & 83.7 & 82.1 & 80.9 & 85.2 & 83.0 & 83.0 & 1.4 \\
    PPSEBM $_{GEN}^{0.2}$ & \textbf{83.4} & \textbf{84.0} & 82.2 & 82.2 & \textbf{85.5} & \textbf{83.3} & \textbf{83.4} & 1.2 \\
    \midrule
    Multitasked & \multicolumn{6}{c}{87.0} \\
  \bottomrule
\end{tabular}}
\label{tab:results}
\end{table*}

\begin{table*}
\caption{Summary of the averaged score on five tasks. The scores are reported as the averaged score over all tasks of the models after training on every task. The rightmost column Multitasked is the upper bound for comparison. The best performance is in boldface.}\centering
    \resizebox{\textwidth}{5mm}{
    \begin{tabular}{ccccccccccc}
        \toprule
        \textbf{Fine-tuned} & \textbf{MAS} & \textbf{LAMOL $_{GEN}^{0.05}$} & \textbf{LAMOL $_{GEN}^{0.2}$} & \textbf{HMI-LAMOL$_{GEN}^{0.05}$} & \textbf{HMI-LAMOL$_{GEN}^{0.2}$} & \textbf{LSEBMCL $_{GEN}^{0.05}$} & \textbf{LSEBMCL $_{GEN}^{0.2}$} & \textbf{PPSEBM $_{GEN}^{0.05}$} & \textbf{PPSEBM $_{GEN}^{0.2}$} & \textbf{Multitasked}\\
        \midrule
        52.6 & 51.2 & 70.3 & 74.1 & 76.0 & 76.9 & 76.5 & 77.3 & \textbf{76.7} & \textbf{77.4} & 78.2 \\ 
        \bottomrule
    \end{tabular}}
\label{tab:fivetasks}
\end{table*}

\begin{table*}
\caption{Summary of results on text classification tasks using averaged EM score (equivalent to averaged accuracy in \cite{de2019episodic}) of models at last epoch of last task. The four orders mirror those in \cite{de2019episodic} as shown in Section \ref{sec:results}, For MBPA++ (our impl.), LAMOL $_{TASK}^{0.2}$, PMR, IDBR, ProgPrompt, HMI-LAMOL, MoCL, SAPT, LSEBMCL$^{0.05}_{GEN}$, and PPSEBM$^{0.05}_{GEN}$, the results are averaged over two runs.}\centering
    \resizebox{\textwidth}{13mm}{
    \begin{tabular}{cccccccccccc}
        \toprule
        \textbf{Order} & \textbf{MBPA++} & \textbf{MBPA++ (our impl.)} & \textbf{LAMOL $_{TASK}^{0.2}$} & \textbf{PMR} & \textbf{IDBR} & \textbf{ProgPrompt} & \textbf{HMI-LAMOL} & \textbf{MoCL} & \textbf{SAPT} & \textbf{LSEBMCL $_{GEN}^{0.05}$} & \textbf{PPSEBM $_{GEN}^{0.05}$} \\
        \midrule
        i & 70.8 & 75.3 & 78.6 & 73.5 & 77.0 & 78.9 & 77.8 & 79.5 & 79.6 & 80.2 & \textbf{80.8}\\ 
        ii & 70.9 & 76.0 & 78.3 & 74.3 & 77.2 & 78.4 & 78.7 & 79.4 & 79.9 & 80.0 & \textbf{80.8}\\
        iii & 70.2 & 73.9 & 78.0 & 72.0 & 78.1 & 79.0 & 79.7 & 79.7 & 79.8 & 80.2 & \textbf{80.5}\\
        iv & 70.7 & 76.7 & 76.9 & 70.2 & 77.8 & 78.0 & 78.4 & 79.5 & 79.5 & \textbf{80.3} & 80.0\\
        \midrule
        Average & 70.7 & 75.5 & 77.9 & 72.5 & 77.5 & 79.0 & 78.6 & 79.5 & 79.7 & 80.2 & \textbf{80.5}\\
        \bottomrule
    \end{tabular}}
\label{tab:classification}
\end{table*}

\subsection{Experimental Setup}

\subsubsection{Tasks, Datasets, and Metrics}
We curated datasets for five diverse natural language processing tasks: question answering, semantic parsing, sentiment analysis, semantic role labeling, and goal-oriented dialogue, as outlined in decaNLP \cite{mccann2018natural}. Additionally, we conducted experiments on four text classification tasks: news classification, sentiment analysis, Wikipedia article classification, and question-and-answer categorization.  Details of the tasks, datasets, and metrics are summarized in Table \ref{tab:data}. The metric scores range from 0 to 100\%. 

\subsubsection{Methods for Comparison}
We compare our proposed approach, PPSEBM, with various existing techniques: (1) \textbf{PPSEBM}: In our experiments, we set $k$ to 1 for top-$k$ sampling and $\lambda_{P}$ to 0.05 representing the weight of the parameter selection loss. We denote PPSEBM with a sampling ratio of $\gamma$ as PPSEBM$^{\gamma}_{GEN}$, where the same GEN token is used across all tasks. (2) \textbf{LAMOL}: Throughout all experiments, LAMOL uses $k = 20$ for top-$k$ sampling and $\lambda = 0.25$ for the weight of the language model (LM) loss. (3) \textbf{RVAE-LAMOL} is the residual variational autoencoder (RVAE) to enhance LAMOL. \cite{wang2022rvae} (4) \textbf{HMI-LAMOL} is the hippocampal memory indexing to enhance the generative replay by controlling sample generation using compressed features of previous training samples \cite{maekawa2023generative}. (5) \textbf{PMR} is an efficient CL method by storing only a few samples to achieve good performance \cite{ho2023prototype}. (6) \textbf{Fine-tuning}: The model is directly fine-tuned on each task sequentially in this baseline. (7) \textbf{Multitask Learning}: This approach trains all tasks simultaneously and is an upper bound for continual learning while ours trains all tasks sequentially. It helps assess whether forgetting is due to a lack of model capacity. (8) \textbf{Regularization-based Methods}: We compare Online EWC \cite{schwarz2018progress} and MAS \cite{aljundi2018memory}, as they are computationally efficient and memory-efficient. (9) \textbf{Gradient Episodic Memory (GEM)}: This approach randomly samples 5\% of the current task size data from the previous task \cite{lopez2017gradient}. (10) \textbf{Improved Memory-Based Parameter Adaptation (MBPA++)}: This method uses sparse experience replay and local adaptation for continual learning, as proposed in \cite{de2019episodic}. (11) \textbf{IDBR}: This approach is an information disentanglement-based regularization method for continual learning on text classification \cite{huang2021continual}. (12) \textbf{ProgPrompt}: This method allows forward transfer and resists catastrophic forgetting \cite{razdaibiedina2023progressive}. (13) \textbf{MoCL}: a rehearsal-free Modular and
Compositional Continual Learning framework
which continually adds new modules to language models and composes them with existing
modules \cite{wang2024rehearsal}. (14) \textbf{SAPT}: a novel
Shared Attention Framework (SAPT), to align
the PET learning and selection via the Shared
Attentive Learning and Selection module \cite{zhao2024sapt}. (15) \textbf{LSEBMCL}: to use energy-based models (EBMs) to prevent catastrophic forgetting by sampling data points from previous tasks when training on new ones \cite{li2025lsebmcl}.

\subsection{Experimental Results}
\label{sec:results}
\begin{figure*}[ht]
\centering
\includegraphics[width=7in]{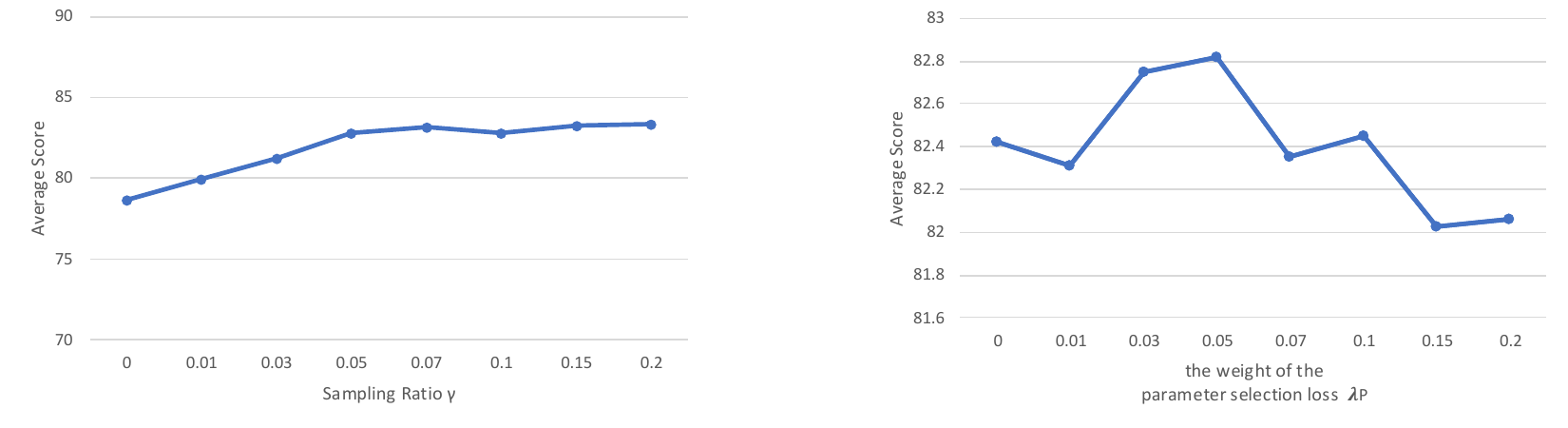}
\caption{Comparison of Different Parameters. Left: sampling ratio $\gamma$. Right: $\lambda_{P}$ for the weight of the parameter selection loss.}
\label{fig:parameter}
\end{figure*}

\subsubsection{SST, QA-SRL, and WOZ Tasks}
To assess the effectiveness of the different methods and the impact of the task order, we conducted experiments on three small datasets: SST, QA-SRL, and WOZ. All methods, except for the multitasked approach, were trained on six different task orders. The model's final score after training on each order was evaluated, and the results are summarized in Table \ref{tab:results}. Several observations were made based on the results: (1) Despite setting $\gamma = 0$, Fine-tuned, EWC, MAS, GEM, LAMOL, and RVAE-LAMOL consistently demonstrate inferior performance compared to PPSEBM. The results further reveal significantly worse performance compared to PPSEBM with $\gamma > 0$. (2) PPSEBM$^{0.2}_{GEN}$ achieves the best overall performance, even approximating the multitasked upper bound with 1.5\%. This suggests that PPSEBM demonstrates minimal forgetting in the context of continual learning. (3) Task order does influence performance with PPSEBM. (4) Performance on old tasks remained consistent throughout the training process with PPSEBM. Increasing the sampling ratio $\gamma$ improved performance, especially when increased from 0 to 0.05. (5) PPSEBM exhibited a low standard deviation among all baselines, indicating its robustness to task order variations.

\subsubsection{Five DecaNLP Tasks}
In this experiment, five tasks (SQuAD 2.0, WikiSQL, SST, QA-SRL, and WOZ) were trained sequentially, starting from the largest task to the smallest one due to limited computing resources. PPSEBM outperformed all other methods (Fine-tuned, MAS, LAMOL$^{0.05}_{GEN}$, LAMOL$^{0.2}_{GEN}$, HMI-LAMOL$^{0.05}_{GEN}$, HMI-LAMOL$^{0.2}_{GEN}$, LSEBMCL $_{GEN}^{0.05}$, and LSEBMCL $_{GEN}^{0.2}$) by a significant margin and even close to the multitasked upper bound with 0.8\%, as shown in Table \ref{tab:fivetasks}. Moreover, the performance of PPSEBM improved with an increase in the sampling ratio $\gamma$.

\subsubsection{Text Classification Tasks}
For text classification tasks, we compared PPSEBM with state-of-the-art methods, MBPA++ \cite{de2019episodic}, LAMOL \cite{sun2019lamol}, IDBR \cite{huang2021continual}, ProgPrompt \cite{razdaibiedina2023progressive}, MoCL \cite{wang2024rehearsal}, SAPT \cite{zhao2024sapt}, and LSEBMCL \cite{li2025lsebmcl} using four different task orders. The results in Table \ref{tab:classification} demonstrated that PPSEBM$_{GEN}^{0.05}$ outperformed LSEBMCL, SAPT, MoCL, ProgPrompt, IDBR, LAMOL$^{0.2}_{TASK}$ and the implementation of MBPA++ even with a sampling ratio of 0.05 and a GEN token. This suggests that the improvements made in PPSEBM were significant and that it is a robust method for mitigating catastrophic forgetting with less sampling data.
We use the dataset orders the same way as \cite{de2019episodic}.

\begin{table*}
\caption{Comparison between GPT2 and large language model Mistral 7B for three tasks. Summary of averaged metric scores for different methods under permuted task orders using models at the last epoch of the last task. The methods without specific notations use the GPT2 pre-trained model. The Average and Std columns respectively are the average and standard deviation of the averaged scores for each row of the methods. Multitasked learning with GPT2 as an upper bound is shown at the bottom.}
  \resizebox{\textwidth}{37mm}{
  \begin{tabular}{lcccccccc}
    \toprule
    \textbf{Model} & \tabincell{c}{\textbf{SST SRL WOZ}} &
    \tabincell{c}{\textbf{SST WOZ SRL}} &
    \tabincell{c}{\textbf{SRL SST WOZ}} & \tabincell{c}{\textbf{SRL W SST}} & 
    \tabincell{c}{\textbf{WOZ SST SRL}} &
    \tabincell{c}{\textbf{WOZ SRL SST}} & \tabincell{c}{\textbf{Average}} & \tabincell{c}{\textbf{Std}} \\ 
    \midrule
    Fine-tuned (GPT2) & 50.2 & 24.7 & 62.9 & 31.3 & 32.8 & 33.9 & 39.3 & 12 \\
    EWC (GPT2) & 50.6 & 48.4 & 64.7 & 35.5 & 43.9 & 39.0 & 47.0 & 8.7 \\
    MAS (GPT2) & 36.5 & 45.3 & 56.6 & 31.0 & 49.7 & 30.8 & 41.6 & 8.9 \\
    GEM (GPT2) & 50.4 & 29.8 & 63.3 & 32.6 & 44.1 & 36.3 & 42.8 & 11 \\
    LAMOL $_{GEN}^{0}$ (GPT2) & 46.5 & 36.6 & 56.6 & 38.6 & 44.9 & 45.2 & 44.8 & 6.0 \\
    LAMOL $_{GEN}^{0.05}$ (GPT2) & 79.6 & 78.9 & 73.1 & 73.7 & 68.6 & 75.7 & 74.9 & 3.4 \\
    LAMOL $_{GEN}^{0.2}$ (GPT2) & 80.0 & 80.7 & 79.6 & 78.7 & 78.4 & 80.5 & 79.7 & 0.8 \\
    RVAE-LAMOL $_{GEN}^{0.05}$ (GPT2) & 79.6 & 79.2 & 79.0 & 78.2 & 77.6 & 79.9 & 78.9 & 0.8 \\
    RVAE-LAMOL $_{GEN}^{0.2}$ (GPT2) & 80.9 & 81.9 & 80.5 & 81.0 & 80.4 & 80.9 & 80.9 & \textbf{0.5} \\
    LSEBMCL $_{GEN}^{0}$ (GPT2) & 65.1 & 64.8 & 66.3 & 66.7 & 69.5 & 65.3 & 66.3 & 1.7 \\
    LSEBMCL $_{GEN}^{0.05}$ (GPT2) & 80.6 & 80.3 & 79.0 & 78.6 & 79.9 & 79.6 & 79.7 & 0.8 \\
    LSEBMCL $_{GEN}^{0.2}$ (GPT2) & 81.1 & 81.5 & 80.4 & 80.7 & 81.7 & 80.6 & 81.0 & \textbf{0.5} \\
    PPSEBM $_{GEN}^{0}$ (GPT2) & 65.3 & 67.8 & 64.2 & 60.6 & 69.3 & 63.0 & 65.0 & 3.2 \\
    PPSEBM $_{GEN}^{0.05}$ (GPT2) & 80.5 & 80.7 & 79.3 & 80.0 & 80.4 & 79.6 & 80.1 & \textbf{0.5} \\
    PPSEBM $_{GEN}^{0.2}$ (GPT2) & 81.5 & 82.0 & 80.4 & 80.4 & 81.7 & 80.7 & 81.1 & 0.7 \\
    PPSEBM $_{GEN}^{0}$ (Mistral) & 78.7 & 54.6 & 75.6 & 62.8 & 61.4 & 61.6 & 65.8 & 9.3 \\
    PPSEBM $_{GEN}^{0.05}$ (Mistral) & 82.8 & 83.7 & 82.1 & 80.9 & 85.2 & 83.0 & 83.0 & 1.4 \\
    PPSEBM $_{GEN}^{0.2}$ (Mistral) & \textbf{83.4} & \textbf{84.0} & \textbf{82.2} & \textbf{82.2} & \textbf{85.5} & \textbf{83.3} & \textbf{83.4} & 1.2 \\
    \midrule
    Multitasked (GPT2) & \multicolumn{6}{c}{82.5} \\
  \bottomrule
\end{tabular}}
\label{tab:results_gpt2}
\end{table*}

\subsection{Comparison with GPT2 Results}
In addition to utilizing the Mistral 7B large language model, we conducted experiments with GPT2. The performance comparison in Table \ref{tab:results_gpt2} indicates that PPSEBM outperforms other baselines even with different pre-trained models, and when coupled with Mistral, outperforms all tasks across various sampling percentages.

\subsection{Forgetting Analysis for Three Tasks}
\begin{figure*}
    \centering
    \includegraphics[width=7in]{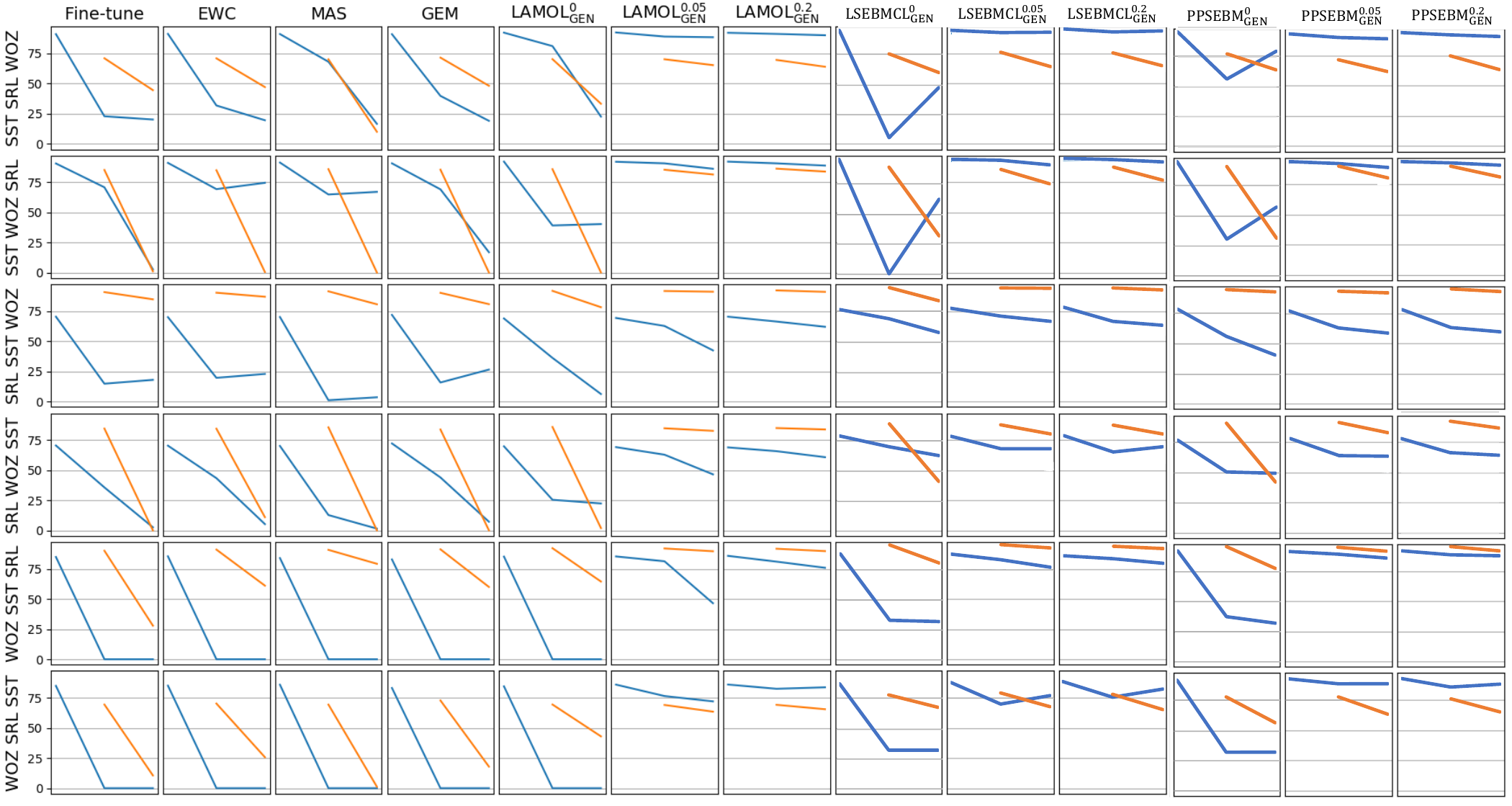}
    \caption{Overview of the forgetting progress for different methods and permuted orders. The blue line indicates the scores of the first task after training each task. The orange line corresponds to that of the second task.}
    \label{fig:Forgetting}
\end{figure*}

Forgetting is analyzed by assessing the performance decrease following each new task, as illustrated in Figure \ref{fig:Forgetting}. The figure highlights that our model consistently exhibits the lowest level of forgetting in most cases, demonstrating the efficiency of the proposed PPSEBM model on reducing forgetting.

\subsection{Parameter Analysis}
In this section, we conduct a comprehensive analysis of two essential parameters: the sampling ratio $\gamma$ and the parameter selction loss weight $\lambda_{P}$. We explore the impact of different sampling ratios: 0, 0.01, 0.03, 0.05, 0.07, 0.1, 0.15, and 0.2. The analysis results are presented in Figure \ref{fig:parameter}. It is evident that increasing the sampling ratio has a substantial positive effect on the average score, particularly during the transition from 0 to 0.05. This indicates that a higher sampling ratio facilitates the generation of diverse and valuable selected parameters, resulting in improved performance in mitigating catastrophic forgetting. We also investigate various values of $\lambda_{P}$, including 0, 0.01, 0.03, 0.05, 0.07, 0.1, 0.15, and 0.2. As $\lambda_{P}$ increases, the average score initially increases and then decreases, achieving the best performance at 0.05. The reason for the oscillation is the randomness of the parameter selection. 

\subsection{Ablation Study}
\begin{table}
\caption{Ablation study. The sequence order is SST, SRL, and WOZ.}
\centering
    \begin{tabular}{lc}
         \toprule
         \textbf{Model} & \textbf{SST SRL WOZ} \\
         \midrule
         No EBM, No PPS & 48.2 \\
         Only EBM & 82.4 \\
         Only PPS & 78.7 \\
         PPSEBM & 83.4 \\
         \bottomrule
    \end{tabular}
\label{tab:ablation}
\end{table}
In this ablation study, we investigate the individual contributions of the Energy-Based Model (EBM) and Progressive Parameter Selection (PPS) components in our proposed PPSEBM method. We conduct two sets of experiments: first, by comparing PPSEBM with and without EBM-generated extra training samples from previous tasks (achieved by setting the sampling ratio to 0), we assess the impact of the EBM in generating informative training data. Second, by setting the parameter selection loss weight $\lambda_{P}$ to 0, we analyze the role of PPS in preserving task-specific selected parameters and mitigating catastrophic forgetting. Table \ref{tab:ablation} presents the results of the ablation study. Without both EBM and PPS, the model achieves the worst performance at 48.2. By including EBM alone, the model's performance significantly improves to 82.4, indicating its importance. Furthermore, only with PPS, the improvement from 48.2 to 78.7 is relatively smaller, affirming that EBM plays a more crucial role in our model. Ultimately, the best performance of 83.4 is achieved when both EBM and PPS are combined. Other datasets also achieved similar results.

\subsection{Complexity Analysis and Implimentation Details}
Referring to Algorithm \ref{alg:PPSEBM}, we observe that in each iteration, we traverse the entire dataset, requiring $O(n)$ time per iteration. Throughout the entire process, we perform $t$ iterations. Therefore, the overall computational complexity amounts to $O(n * t)$. In our tests, the sampling took less than 1 minute per 0.05 sampling ratio after finishing training the energy-based model. The total training time of EBM is 17 minutes for three tasks (SST, SRL, WOZ). The training time for the entire model, excluding EBM on the SST, SRL, WOZ tasks with the same ratio was approximately 19 hours, where the individual SST task took 7 hours. More specifically, in the architecture of EBM, we use an encoder-decoder framework. The encoder has 8192 * 512 * 2 = 8M parameters and the decoder has 8192 * 512 + 512 * 32011 = 20M parameters, a total 28M parameters, which is significantly less than the LLM Mistral 7B which has 7B (250 times of EBM) parameters and GPT2 which has 117M (4 times of EBM) parameters. This is why the cost of EBM is insignificant compared to the main training process. Our experimental setup includes basic settings with $\lambda_{P} = 0.05$ and $\gamma = 0.05$. To optimize GPU usage and expedite training, we implement quantization methods to reduce model parameters. Notably, our experiments use a single NVIDIA Corporation TU102GL [Quadro RTX 6000/8000] 48 GB GPU per experiment setting.

\begin{table}
\caption{Summary of averaged metric scores for different methods under SelfRC, TweetQA, and SST task order using models at the last epoch of the last task. Multitasked learning as an upper bound is shown at the bottom.}\centering
  \begin{tabular}{lc}
    \toprule
    \textbf{Model} & \tabincell{c}{\textbf{SelfRC TweetQA SST}}\\ 
    \midrule
    Fine-tuned & 49.8 \\
    EWC & 49.0 \\
    MAS & 38.9 \\
    GEM & 50.0 \\
    LAMOL $_{GEN}^{0}$ & 45.2 \\
    LAMOL $_{GEN}^{0.05}$ & 76.0 \\
    LAMOL $_{GEN}^{0.2}$ & 76.3 \\
    RVAE-LAMOL $_{GEN}^{0.05}$ & 76.1 \\
    RVAE-LAMOL $_{GEN}^{0.2}$ & 77.2 \\
    LSEBMCL $_{GEN}^{0}$ & 55.4 \\
    LSEBMCL $_{GEN}^{0.05}$ & 78.9 \\
    LSEBMCL $_{GEN}^{0.2}$ & 80.2 \\
    PPSEBM $_{GEN}^{0}$ & 57.6 \\
    PPSEBM $_{GEN}^{0.05}$ & 79.8 \\
    PPSEBM $_{GEN}^{0.2}$ & \textbf{80.9} \\
    \midrule
    Multitasked & 82.8 \\
    \bottomrule
    \end{tabular}
    \label{tab:recent_results}
\end{table}

\subsection{Results for SelfRc, TweetQA, and SST}
We also do experiments with the most recent SQuAD format datasets SelfRC \cite{saha2018duorc}, TweetQA \cite{xiong2019tweetqa}, together with the dataset SST \cite{radford2017learning}. The comparison results are shown in Table \ref{tab:recent_results}. From the results, we can see we not only perform best in older datasets but also achieve the best performance and approximates the upper bound, Multitasked with 1.9\%.

\section{Conclusion}
In this work, we proposed PPSEBM, a continual learning framework that effectively addresses catastrophic forgetting by integrating Progressive Parameter Selection (PPS) with an Energy-Based Model (EBM)–based generative replay. Our experiments demonstrate that PPSEBM achieves superior performance over state-of-the-art baselines across diverse natural language processing tasks. A central strength of PPSEBM lies in the tight coupling between its components: the EBM not only generates representative samples from previous tasks but also directly informs and guides the PPS process, shaping which parameters are prioritized for updating. This dynamic interaction enables the system to balance knowledge retention and adaptability, allowing PPS to make more targeted and effective updates. Looking ahead, we plan to explore deeper integration between EBM and parameter selection strategies and to investigate the incorporation of alternative generative models, such as diffusion models.

\section*{Acknowledgement}
The research reported herein was supported in part by NIST under grant number 60NANB24D143, and by the National Center for Transportation Cybersecurity and Resiliency (TraCR) (a U.S. Department of Transportation National University Transportation Center) headquartered at Clemson University, Clemson, South Carolina, USA, under grant number 69A3552344812. Any opinions, findings, conclusions, and recommendations expressed in this material are those of the author(s) and do not necessarily reflect the views of NIST or TraCR. The U.S. Government assumes no liability for the contents or use thereof.

\bibliographystyle{IEEEtran}
\bibliography{IEEEabrv}

@article{shin2017continual,
  title={Continual learning with deep generative replay},
  author={Shin, Hanul and Lee, Jung Kwon and Kim, Jaehong and Kim, Jiwon},
  journal={arXiv preprint arXiv:1705.08690},
  year={2017}
}

@inproceedings{aljundi2018memory,
  title={Memory aware synapses: Learning what (not) to forget},
  author={Aljundi, Rahaf and Babiloni, Francesca and Elhoseiny, Mohamed and Rohrbach, Marcus and Tuytelaars, Tinne},
  booktitle={Proceedings of the European Conference on Computer Vision (ECCV)},
  pages={139--154},
  year={2018}
}

@article{lopez2017gradient,
  title={Gradient episodic memory for continual learning},
  author={Lopez-Paz, David and Ranzato, Marc'Aurelio},
  journal={Advances in neural information processing systems},
  volume={30},
  pages={6467--6476},
  year={2017}
}

@article{chaudhry2018efficient,
  title={Efficient lifelong learning with a-gem},
  author={Chaudhry, Arslan and Ranzato, Marc'Aurelio and Rohrbach, Marcus and Elhoseiny, Mohamed},
  journal={arXiv preprint arXiv:1812.00420},
  year={2018}
}

@article{li2017learning,
  title={Learning without forgetting},
  author={Li, Zhizhong and Hoiem, Derek},
  journal={IEEE transactions on pattern analysis and machine intelligence},
  volume={40},
  number={12},
  pages={2935--2947},
  year={2017},
  publisher={IEEE}
}

@article{lecun2006tutorial,
  title={A tutorial on energy-based learning},
  author={LeCun, Yann and Chopra, Sumit and Hadsell, Raia and Ranzato, M and Huang, Fujie},
  journal={Predicting structured data},
  volume={1},
  number={0},
  year={2006}
}

@article{du2019implicit,
  title={Implicit generation and generalization in energy-based models},
  author={Du, Yilun and Mordatch, Igor},
  journal={arXiv preprint arXiv:1903.08689},
  year={2019}
}

@inproceedings{xie2016theory,
  title={A theory of generative convnet},
  author={Xie, Jianwen and Lu, Yang and Zhu, Song-Chun and Wu, Yingnian},
  booktitle={International Conference on Machine Learning},
  pages={2635--2644},
  year={2016},
  organization={PMLR}
}

@article{nijkamp2019learning,
  title={Learning non-convergent non-persistent short-run MCMC toward energy-based model},
  author={Nijkamp, Erik and Hill, Mitch and Zhu, Song-Chun and Wu, Ying Nian},
  journal={Advances in Neural Information Processing Systems},
  volume={32},
  year={2019}
}

@article{pang2020learning,
  title={Learning latent space energy-based prior model},
  author={Pang, Bo and Han, Tian and Nijkamp, Erik and Zhu, Song-Chun and Wu, Ying Nian},
  journal={Advances in Neural Information Processing Systems},
  volume={33},
  pages={21994--22008},
  year={2020}
}

@article{zhang2021learning,
  title={Learning generative vision transformer with energy-based latent space for saliency prediction},
  author={Zhang, Jing and Xie, Jianwen and Barnes, Nick and Li, Ping},
  journal={Advances in Neural Information Processing Systems},
  volume={34},
  pages={15448--15463},
  year={2021}
}

@inproceedings{li2022energy,
  title={Energy-based models for continual learning},
  author={Li, Shuang and Du, Yilun and van de Ven, Gido and Mordatch, Igor},
  booktitle={Conference on Lifelong Learning Agents},
  pages={1--22},
  year={2022},
  organization={PMLR}
}

@article{mccann2018natural,
  title={The natural language decathlon: Multitask learning as question answering},
  author={McCann, Bryan and Keskar, Nitish Shirish and Xiong, Caiming and Socher, Richard},
  journal={arXiv preprint arXiv:1806.08730},
  year={2018}
}

@inproceedings{li2022lpc,
  title={LPC: A Logits and Parameter Calibration Framework for Continual Learning},
  author={Li, Xiaodi and Wang, Zhuoyi and Li, Dingcheng and Khan, Latifur and Thuraisingham, Bhavani},
  booktitle={Findings of the Association for Computational Linguistics: EMNLP 2022},
  pages={7142--7155},
  year={2022}
}

@article{de2019episodic,
  title={Episodic memory in lifelong language learning},
  author={de Masson D'Autume, Cyprien and Ruder, Sebastian and Kong, Lingpeng and Yogatama, Dani},
  journal={Advances in Neural Information Processing Systems},
  volume={32},
  year={2019}
}

@inproceedings{schwarz2018progress,
  title={Progress \& compress: A scalable framework for continual learning},
  author={Schwarz, Jonathan and Czarnecki, Wojciech and Luketina, Jelena and Grabska-Barwinska, Agnieszka and Teh, Yee Whye and Pascanu, Razvan and Hadsell, Raia},
  booktitle={International conference on machine learning},
  pages={4528--4537},
  year={2018},
  organization={PMLR}
}

@article{sun2019lamol,
  title={Lamol: Language modeling for lifelong language learning},
  author={Sun, Fan-Keng and Ho, Cheng-Hao and Lee, Hung-Yi},
  journal={arXiv preprint arXiv:1909.03329},
  year={2019}
}

@article{radford2017learning,
  title={Learning to generate reviews and discovering sentiment},
  author={Radford, Alec and Jozefowicz, Rafal and Sutskever, Ilya},
  journal={arXiv preprint arXiv:1704.01444},
  year={2017}
}

@article{lester2021power,
  title={The power of scale for parameter-efficient prompt tuning},
  author={Lester, Brian and Al-Rfou, Rami and Constant, Noah},
  journal={arXiv preprint arXiv:2104.08691},
  year={2021}
}

@article{vu2021spot,
  title={Spot: Better frozen model adaptation through soft prompt transfer},
  author={Vu, Tu and Lester, Brian and Constant, Noah and Al-Rfou, Rami and Cer, Daniel},
  journal={arXiv preprint arXiv:2110.07904},
  year={2021}
}

@article{huang2021continual,
  title={Continual learning for text classification with information disentanglement based regularization},
  author={Huang, Yufan and Zhang, Yanzhe and Chen, Jiaao and Wang, Xuezhi and Yang, Diyi},
  journal={arXiv preprint arXiv:2104.05489},
  year={2021}
}

@article{hinton2006reducing,
  title={Reducing the dimensionality of data with neural networks},
  author={Hinton, Geoffrey E and Salakhutdinov, Ruslan R},
  journal={science},
  volume={313},
  number={5786},
  pages={504--507},
  year={2006},
  publisher={American Association for the Advancement of Science}
}

@article{bengio2009learning,
  title={Learning deep architectures for AI},
  author={Bengio, Yoshua and others},
  journal={Foundations and trends{\textregistered} in Machine Learning},
  volume={2},
  number={1},
  pages={1--127},
  year={2009},
  publisher={Now Publishers, Inc.}
}

@article{lecun2015deep,
  title={Deep learning},
  author={LeCun, Yann and Bengio, Yoshua and Hinton, Geoffrey},
  journal={nature},
  volume={521},
  number={7553},
  pages={436--444},
  year={2015},
  publisher={Nature Publishing Group UK London}
}

@article{brown2020language,
  title={Language models are few-shot learners},
  author={Brown, Tom and Mann, Benjamin and Ryder, Nick and Subbiah, Melanie and Kaplan, Jared D and Dhariwal, Prafulla and Neelakantan, Arvind and Shyam, Pranav and Sastry, Girish and Askell, Amanda and others},
  journal={Advances in neural information processing systems},
  volume={33},
  pages={1877--1901},
  year={2020}
}

@article{holtzman2019curious,
  title={The curious case of neural text degeneration},
  author={Holtzman, Ari and Buys, Jan and Du, Li and Forbes, Maxwell and Choi, Yejin},
  journal={arXiv preprint arXiv:1904.09751},
  year={2019}
}

@article{petroni2019language,
  title={Language models as knowledge bases?},
  author={Petroni, Fabio and Rockt{\"a}schel, Tim and Lewis, Patrick and Bakhtin, Anton and Wu, Yuxiang and Miller, Alexander H and Riedel, Sebastian},
  journal={arXiv preprint arXiv:1909.01066},
  year={2019}
}

@article{zhao2024sapt,
  title={Progressive prompts: Continual learning for language models},
  author={Razdaibiedina, Anastasia and Mao, Yuning and Hou, Rui and Khabsa, Madian and Lewis, Mike and Almahairi, Amjad},
  journal={arXiv preprint arXiv:2301.12314},
  year={2023}
}

@article{jiang2023mistral,
  title={Mistral 7B},
  author={Jiang, Albert Q and Sablayrolles, Alexandre and Mensch, Arthur and Bamford, Chris and Chaplot, Devendra Singh and Casas, Diego de las and Bressand, Florian and Lengyel, Gianna and Lample, Guillaume and Saulnier, Lucile and others},
  journal={arXiv preprint arXiv:2310.06825},
  year={2023}
}

@inproceedings{wang2022rvae,
  title={RVAE-LAMOL: Residual Variational Autoencoder to Enhance Lifelong Language Learning},
  author={Wang, Han and Fu, Ruiliu and Zhang, Xuejun and Zhou, Jun},
  booktitle={2022 International Joint Conference on Neural Networks (IJCNN)},
  pages={1--9},
  year={2022},
  organization={IEEE}
}

@inproceedings{maekawa2023generative,
  title={Generative Replay Inspired by Hippocampal Memory Indexing for Continual Language Learning},
  author={Maekawa, Aru and Kamigaito, Hidetaka and Funakoshi, Kotaro and Okumura, Manabu},
  booktitle={Proceedings of the 17th Conference of the European Chapter of the Association for Computational Linguistics},
  pages={930--942},
  year={2023}
}

@article{ho2023prototype,
  title={Prototype-guided memory replay for continual learning},
  author={Ho, Stella and Liu, Ming and Du, Lan and Gao, Longxiang and Xiang, Yong},
  journal={IEEE Transactions on Neural Networks and Learning Systems},
  year={2023},
  publisher={IEEE}
}

@article{saha2018duorc,
  title={DuoRC: Towards complex language understanding with paraphrased reading comprehension},
  author={Saha, Amrita and Aralikatte, Rahul and Khapra, Mitesh M and Sankaranarayanan, Karthik},
  journal={arXiv preprint arXiv:1804.07927},
  year={2018}
}

@article{xiong2019tweetqa,
  title={TWEETQA: A social media focused question answering dataset},
  author={Xiong, Wenhan and Wu, Jiawei and Wang, Hong and Kulkarni, Vivek and Yu, Mo and Chang, Shiyu and Guo, Xiaoxiao and Wang, William Yang},
  journal={arXiv preprint arXiv:1907.06292},
  year={2019}
}

@article{zhou2023revisiting,
  title={Revisiting class-incremental learning with pre-trained models: Generalizability and adaptivity are all you need},
  author={Zhou, Da-Wei and Cai, Zi-Wen and Ye, Han-Jia and Zhan, De-Chuan and Liu, Ziwei},
  journal={arXiv preprint arXiv:2303.07338},
  year={2023}
}

@inproceedings{jung2023generating,
  title={Generating instance-level prompts for rehearsal-free continual learning},
  author={Jung, Dahuin and Han, Dongyoon and Bang, Jihwan and Song, Hwanjun},
  booktitle={Proceedings of the IEEE/CVF International Conference on Computer Vision},
  pages={11847--11857},
  year={2023}
}

@inproceedings{kim2025open,
  title={Open-world dynamic prompt and continual visual representation learning},
  author={Kim, Youngeun and Fang, Jun and Zhang, Qin and Cai, Zhaowei and Shen, Yantao and Duggal, Rahul and S Raychaudhuri, Dripta and Tu, Zhuowen and Xing, Yifan and Dabeer, Onkar},
  booktitle={European Conference on Computer Vision},
  pages={357--374},
  year={2025},
  organization={Springer}
}

@article{zhou2024continual,
  title={Continual learning with pre-trained models: A survey},
  author={Zhou, Da-Wei and Sun, Hai-Long and Ning, Jingyi and Ye, Han-Jia and Zhan, De-Chuan},
  journal={arXiv preprint arXiv:2401.16386},
  year={2024}
}

@inproceedings{li2025lsebmcl,
  title={LSEBMCL: A Latent Space Energy-Based Model for Continual Learning},
  author={Li, Xiaodi and Li, Dingcheng and Gao, Rujun and Zamani, Mahmoud and Khan, Latifur},
  booktitle={2025 International Conference on Artificial Intelligence in Information and Communication (ICAIIC)},
  pages={0690--0695},
  year={2025},
  organization={IEEE}
}

@article{wang2024rehearsal,
  title={Rehearsal-free modular and compositional continual learning for language models},
  author={Wang, Mingyang and Adel, Heike and Lange, Lukas and Str{\"o}tgen, Jannik and Sch{\"u}tze, Hinrich},
  journal={arXiv preprint arXiv:2404.00790},
  year={2024}
}

@inproceedings{wang2022dualprompt,
  title={Dualprompt: Complementary prompting for rehearsal-free continual learning},
  author={Wang, Zifeng and Zhang, Zizhao and Ebrahimi, Sayna and Sun, Ruoxi and Zhang, Han and Lee, Chen-Yu and Ren, Xiaoqi and Su, Guolong and Perot, Vincent and Dy, Jennifer and others},
  booktitle={European conference on computer vision},
  pages={631--648},
  year={2022},
  organization={Springer}
}

@article{razdaibiedina2023progressive,
  title={Progressive prompts: Continual learning for language models},
  author={Razdaibiedina, Anastasia and Mao, Yuning and Hou, Rui and Khabsa, Madian and Lewis, Mike and Almahairi, Amjad},
  journal={arXiv preprint arXiv:2301.12314},
  year={2023}
}

@inproceedings{wang2022learning,
  title={Learning to prompt for continual learning},
  author={Wang, Zifeng and Zhang, Zizhao and Lee, Chen-Yu and Zhang, Han and Sun, Ruoxi and Ren, Xiaoqi and Su, Guolong and Perot, Vincent and Dy, Jennifer and Pfister, Tomas},
  booktitle={Proceedings of the IEEE/CVF conference on computer vision and pattern recognition},
  pages={139--149},
  year={2022}
}

\end{document}